\documentclass[runningheads]{llncs}
\usepackage{graphicx}
\usepackage{array} 
\usepackage{amsmath}
\usepackage{amssymb}
\usepackage{algorithm}
\usepackage{algorithmic}
\usepackage{bm}
\usepackage{cases} 
\usepackage{subfigure}
\usepackage{graphicx}
\usepackage{amstext}
\usepackage{dsfont}
\usepackage{fixmath} 
\usepackage{units}
\usepackage{booktabs}
\usepackage{color}

\newcommand{\mB}[1]{\mathbf{#1}}

\begin{document}
\title{Bidirectional Modeling and Analysis of Brain Aging with Normalizing Flows}

\titlerunning{Bidirectional Modeling and Analysis of Brain Aging with Normalizing Flows}
%
\author{Matthias Wilms$^{1,2,3}$ \and Jordan J. Bannister$^{1,2,3}$ \and Pauline Mouches$^{1,2,3}$ \and M. Ethan MacDonald$^{1,2}$ \and Deepthi Rajashekar$^{1,2,3}$\and Sönke Langner$^4$ \and Nils D. Forkert$^{1,2,3}$}

\authorrunning{Wilms et.\,al.}
%
\institute{Department of Radiology, University of Calgary, Canada \and
Hotchkiss Brain Institute, University of Calgary, Canada\and
Alberta Children’s Hospital Research Institute, University of Calgary, Canada\and
Institute for Diagnostic and Interventional Radiology, Pediatric and Neuroradiology, University Medical Center Rostock, Germany\\
\email{matthias.wilms@ucalgary.ca}}
\maketitle              
\begin{abstract}
Brain aging is a widely studied longitudinal process throughout which the brain undergoes considerable morphological changes and various machine learning approaches have been proposed to analyze it. Within this context, brain age prediction from structural MR images and age-specific brain morphology template generation are two problems that have attracted much attention. While most approaches tackle these tasks independently, we assume that they are inverse directions of the same functional bidirectional relationship between a brain's morphology and an age variable. In this paper, we propose to model this relationship with a single conditional normalizing flow, which unifies brain age prediction and age-conditioned generative modeling in a novel way. In an initial evaluation of this idea, we show that our normalizing flow brain aging model can accurately predict brain age while also being able to generate age-specific brain morphology templates that realistically represent the typical aging trend in a healthy population. This work is a step towards unified modeling of functional relationships between 3D brain morphology and clinical variables of interest with powerful normalizing flows.
\end{abstract}
\section{Introduction}\label{sec:intro}
Many machine learning (ML) tasks in neuroimaging aim at modeling and exploring complex functional relationships between brain morphology derived from structural MR images and clinically relevant scores and variables of interest\,\cite{Mateos_NeuroClinical_2018}. In this context, the aging process of the brain throughout which it undergoes considerable morphological changes is a widely studied example. Many prediction models have been proposed to estimate a brain's biological age from a patient's structural MRI data\,(see overview in \cite{Cole_Book_2019}). The potential difference between a patient's predicted biological brain age and the true chronological age is an early indicator for neurodegenerative disorders like Alzheimer's disease\,\cite{Cole_2019}. 

\begin{figure}[t]
  \centering		    
	\includegraphics[width=\textwidth]{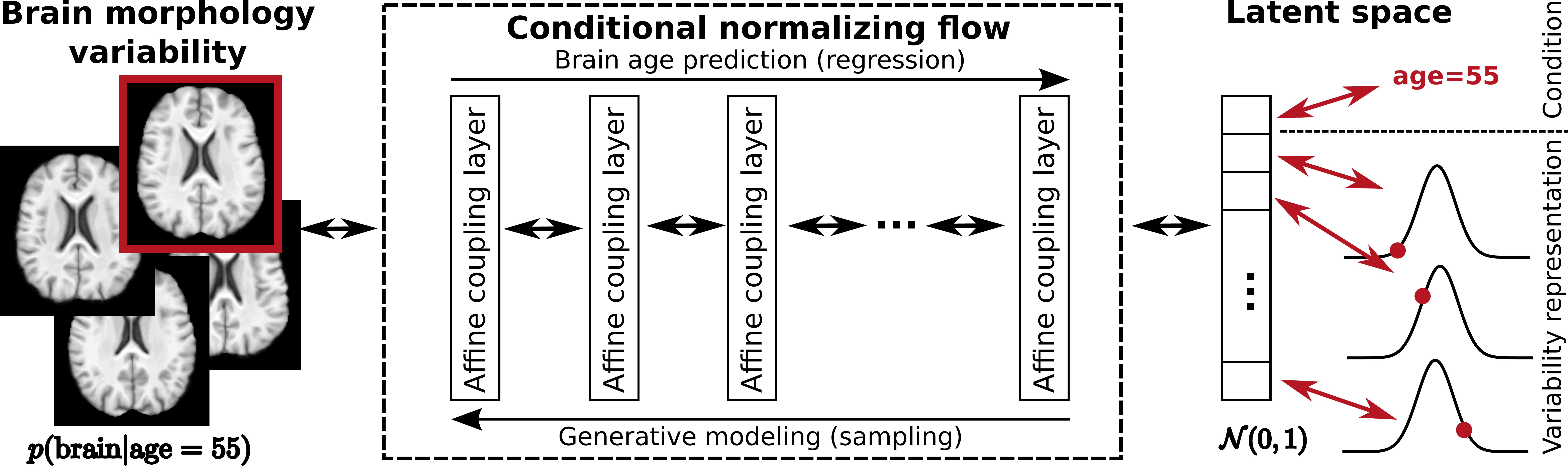}
	\caption{Graphical overview of the proposed modeling approach based on a conditional normalizing flow composed of affine coupling layers. The bidirectional flow maps brain morphology variability to a structured latent space. It solves the brain age prediction problem (left to right) and can be utilized to sample from the distribution of brain morphology conditioned on age (right to left). The first component of the latent space encodes brain age while all other dimensions represent variability. See text for details.}
	\label{Fig0}
\end{figure}

In contrast, modeling the inverse direction of this classical regression task allows the study of morphological changes associated with a certain age on a population level, which can be useful to assist basic research \cite{Cole_2019}, or for patient-specific brain aging simulation \cite{Xia_MICCAI_2019,Wegmayr_GCPR_2019}. This generative modeling problem is closely related to the numerous atlas (a.k.a. template) building approaches that aim at computing a model of a population's average anatomy conditioned on age \cite{Huizinga_Neuro_2018,Sivera_Neuro_2019}. Several recent papers model the joint distribution of brain morphology and age with deep learning (DL) techniques such as variants of Generative Adversarial Networks~(GANs), Variational Autoencoders~(VAEs), or related concepts \cite{Xia_MICCAI_2019,Wegmayr_GCPR_2019,Dalca_NIPS_2019,Zhao_MICCAI_2019}. However, most of them only focus on single slices or small 3D regions of the brain due to the high computational costs of these DL techniques.

Despite the volumes of literature available on brain age prediction and age-conditioned generative brain modeling, the fact that these tasks constitute inverse directions of the same bidirectional functional relationship between age and morphology is usually ignored, and independent or loosely coupled models for each problem are created. A notable exception is \cite{Zhao_MICCAI_2019}, where a VAE is equipped with a linear regression model that maps the VAE's latent space representation of a brain to its age. This component allows the use of the VAE's encoder for age prediction and the decoder can generate age-conditioned brains. However, this extension still does not guarantee that the encoder and the decoder are inverses of each other, which potentially leads to inconsistencies. Additionally, training a VAE is usually difficult due to intractable likelihoods and posterior collapse.

In this work, we propose to model the bidirectional functional relationship between brain morphology and brain age in a unified, consistent way using conditional normalizing flows. Normalizing flows (NFs) \cite{Kobyzev_PAMI_2020} learn invertible functions between a complex probability distribution of interest and a simple base distribution. In contrast to GANs and VAEs, they can be trained directly via maximum
likelihood estimation, they do not suffer from posterior collapse, sampling is very efficient, and both directions are the exact inverses of each other \cite{Kobyzev_PAMI_2020}. 

Our bidirectional NF-based brain aging model is based on ideas about unifying regression and generative modeling using NFs described in \cite{Ardizzone_ICLR_2019}. Our model (1) learns the distribution of brain morphology conditioned on age of a population, which can be sampled to generate age-conditioned brain templates, and (2) is able to predict a brain's biological age given its structural MR image (see Fig. \ref{Fig0}). Moreover, we propose pre-processing steps to encode morphological variability based on diffeomorphic transformations that allow us to directly handle whole 3D images. To our knowledge, this is the first conditional NF for bidirectional modeling and analysis of brain aging directly utilizing 3D structural MRI data.
\section{Problem Formulation and Pre-Processing}\label{sec:problem}
We will first introduce our notation and the modeling problem, then two pre-processing steps are described in Sec. \ref{sec:problem:diffeo} and Sec. \ref{sec:problem:pca}. These steps help us to efficiently model the brain aging problem described in Sec. \ref{sec:cnf} with a NF.

For our analysis, we assume a training population $\{(I_i,a_i)\}_{i=1}^{n_{\text{pop}}}$ of $n_{\text{pop}}$ healthy subjects\footnote{We assume that for healthy subjects, chronological and biological brain age are equal.} to be given. Each tuple $(I_i,a_i)$ consists of a subject's 3D structural MRI brain scan $I_i:\mathbb{R}^3\to\mathbb{R}$ and the associated chronological brain age $a_i\in\mathbb{R}$. The goal of this work is to train a combined regression and generative NF model that is able (1) to estimate the brain age $a\in\mathbb{R}$ of a new subject with brain scan $I:\mathbb{R}^3\to\mathbb{R}$, and (2) to accurately approximate the distribution of plausible brain morphology conditioned on age.
\subsection{Reference Space and Deformation-based Analysis}\label{sec:problem:diffeo}
Instead of directly using the structural MR images, our method follows the classical concepts from deformation-based morphometry \cite{Ashburner_HBM_1998} to represent morphological differences with respect to a common template via non-linear, diffeomorphic transformations. A deformation-based approach is appropriate here as aging mostly results in shape changes of various brain structures (e.g., cerebral atrophy) and it allows our NF to directly focus on shape differences for regression and we avoid typical problems of generative models (e.g., blurry images).

Following standard practice in deformation-based brain aging modeling \cite{Sivera_Neuro_2019}, a template image $\overline{I}:\mathbb{R}^3\to\mathbb{R}$ with a reference brain morphology serves as a reference space for all further computations. The template can either be computed specifically for the training population or a standard template can also be used. We map all the training images non-linearly to this template via diffeomorphic image registration resulting in $n_{\text{pop}}$ spatial transformations $\{\varphi_i:\mathbb{R}^{3}\to\mathbb{R}^3\}_{i=1}^{n_{\text{pop}}}$, which encode the morphological differences between the template and each subject. Here, each diffeomorphic transformation $\varphi_i=\text{exp}(v_i)$ is parameterized via a stationary velocity field $v_i:\mathbb{R}^3\to\mathbb{R}^3$ where $\text{exp}(\cdot)$ denotes the group exponential map from the Log-Euclidean framework (see \cite{Arsigny_MICCAI_2006} for details).
\subsection{Dimensionality Reduction via PCA on Diffeomorphisms}\label{sec:problem:pca}
Directly utilizing the previously computed diffeomorphic transformations for 3D NF-based brain aging modeling is challenging due to the high dimensionality of the data (MRI brain scans usually consist of several millions of voxels $n_{\text{vox}}$), which is usually also magnitudes larger than the number of training samples $n_{\text{pop}}$. However, it is safe to assume that the transformations, which are vector fields, contain redundant information and noise allowing for dimensionality reduction. Following previous results on brain modeling \cite{Gerber_MedIA_2010,Zhang_MedIA_2017}, we, therefore, assume that a low dimensional structure of the space of brain shapes exists. More specifically, we assume that all plausible transformations needed for aging modeling lie within a $n_{\text{sub}}$-dimensional affine subspace of maximum data variation $\mathcal{A}=\{\overline{\mB{v}}+\mB{q}\ |\  \mB{q}\in\text{span}(\mB{Q})\}\ \text{with}\ n_{\text{sub}}< n_{\text{pop}}$ of the velocity fields of the training data. Here, $\overline{\mB{v}}\in\mathbb{R}^{3n_{\text{vox}}}$ denotes the vectorized mean velocity field of the training data and $\mB{Q}\in\mathbb{R}^{3n_{\text{vox}}\times n_{\text{sub}}}$ is an orthonormal column matrix of the first $n_{\text{sub}}$ principal components resulting from a principal components analysis (PCA) of the velocity fields $\{v_i\}_{i=1}^{n_{\text{pop}}}$ \cite{Arsigny_MICCAI_2006,Ehrhardt_SPIE_2010}. Performing statistics directly on velocity fields will preserve diffeomorphisms and the projection matrix $\mB{Q}$ (with pseudoinverse $\mB{Q}^T$) can be directly integrated into the NF model.
\section{Normalizing Flow Model for Brain Aging Analysis}\label{sec:cnf}
Applying the pre-processing steps detailed in Sec. \ref{sec:problem} to the training data results in a set $\{(\mB{v}_i,a_i)\}_{i=1}^{n_{\text{pop}}}$. Here, $\mB{v}_i\in\mathbb{R}^{n_{sub}}$ denotes a projected velocity field $v_i$ in coordinates of subspace $\mathcal{A}$. Given the training tuples, our goals are two-fold: (1) Learn a function $f(\cdot;\theta)$ with parameters $\theta$ that takes a new morphology-encoding velocity field representation $\mB{v}\in\mathbb{R}^{n_{\text{sub}}}$ and predicts its age $a\in\mathbb{R}$: $a=f(\mB{v};\theta)$. (2) Train a generative model to efficiently sample velocity fields from the conditional distribution of brain morphologies conditioned on age: $p(\mB{v}|a)$.
\subsection{Bidirectional Conditional Modeling}
The innovative idea of this work is to solve both problems with a unified, bidirectional conditional NF model. In general, a NF represents a complex, bijective function between two sets as a chain of simpler sub-functions \cite{Kobyzev_PAMI_2020}. At first, the bijective property seems to be incompatible with our setup as we are mapping from $\mathbb{R}^{n_{\text{sub}}}$ (velocity fields) to $\mathbb{R}$ (age). However, to model $p(\mB{v}|a)$ we also need to encode the (inter-subject) morphological variability associated with brains at the same age. We therefore follow \cite{Ardizzone_ICLR_2019} to combine regression and generative modeling with an NF that is conditioned on the regression target.

Our conditional NF model represents a bijective function $f(\cdot;\theta):\mathbb{R}^{n_{\text{sub}}}\to \mathbb{R}^{n_{\text{sub}}}$ that maps a $n_{\text{sub}}$-dimensional velocity field to a latent space of the same size. Aiming for a structured latent space, we define that one dimension of the latent space accounts for the age $a$ of the input $\mB{v}$, solving the prediction task while conditioning the flow. All other dimensions store the age-unrelated information $\mB{z}\in\mathbb{R}^{n_{\text{sub}}-1}$ needed to reconstruct the input. In the following, sub-parts of $f(\cdot;\theta)$ mapping $\mB{v}$ to $a$ will be named $f_a(\cdot;\theta)$.

Imposing a simple prior on $\mB{z}\sim p(\mB{z})$ (e.g., Gaussian distribution with diagonal covariance) and assuming that the distribution $p(f_a(\mB{v};\theta)|a)$ associated with the age prediction part can be modeled easily and independently (e.g., age-independent Gaussian residuals for squared error) allow us to relate $p(\mB{v}|a)$ and the latent space via $f(\cdot;\theta)$ with the change of variables theorem \cite{Ardizzone_ICLR_2019,Xiao_Arxiv_2019}:
\begin{equation}
    p(\mB{v}|a)=p(\mB{z})p(f_a(\mB{v};\theta)|a)|J|^{-1}\ \text{with}\ J=\text{det}\Bigg(\frac{\partial f^{-1}([a,\mB{z}];\theta)}{\partial [a,\mB{z}]}\Bigg)\ .\label{eq:cov}
\end{equation}
Here, $f^{-1}(\cdot;\theta)$ denotes the inverse of $f(\cdot;\theta)$ and $J$ represents the associated Jacobian determinant. Given an invertible function $f(\cdot;\theta)$, samples from the simple priors can be transformed to approximate $p(\mB{v}|a)$.

\subsection{Normalizing Flow Architecture and Training}\label{sec:nf:architecture}
The challenge resulting from Eq. (\ref{eq:cov}) is to find an easily invertible function $f(\cdot;\theta)$ with a tractable Jacobian. In NFs, this is done by first defining $[a,\mB{z}]=f(\mB{v};\theta)=f_{n_{\text{lay}}}\circ\cdots\circ f_i \circ\cdots\circ f_1(\mB{v})$ as a chain of $n_{\text{lay}}$ simpler, invertible sub-functions $f_i(\cdot,\theta_i)$ (also named coupling layers). Our NF model consists of affine coupling layers \cite{Dinh_ICLR_2017,Ardizzone_ICLR_2019}, which are a common choice in previous NF research due to their flexibility and favourable computational properties \cite{Kobyzev_PAMI_2020}. 

Let $\mB{u}=[\mB{u}_1,\mB{u}_2]\in\mathbb{R}^{n_{\text{sub}}}$ and $\mB{w}=[\mB{w}_1,\mB{w}_2]\in\mathbb{R}^{n_{\text{sub}}}$ denote input and output vectors of an affine coupling layer, where $\mB{u}_1$ and $\mB{w}_1$ represent the first $n_{\text{sub}}/2$ dimensions while $\mB{u}_2$ and $\mB{w}_2$ cover the second half. Then, an affine coupling layer $\mB{w}=f_i(\mB{u},\theta_i)$, with $\mB{u}=[\mB{u}_1,\mB{u}_2]\in\mathbb{R}^{n_{\text{sub}}}$ and $\mB{w}=[\mB{w}_1,\mB{w}_2]\in\mathbb{R}^{n_{\text{sub}}}$, defines an element-wise affine transformation parameterized by $\mB{u}_2$ that maps $\mB{u}_1$ to $\mB{w}_1$ \cite{Dinh_ICLR_2017}:
\begin{align}
\mB{w}_1&=&\text{exp}\big(s(\mB{u}_2,\theta_i)\big)\odot\mB{u}_1+t(\mB{u}_2,\theta_i)\ \ \ \text{and}\ \ \ \mB{w}_2&=&\mB{u}_2\ .\nonumber
\end{align}
Here, the scaling function $s(\cdot,\theta_i)$ and the translation function $t(\cdot,\theta_i)$ can be arbitrarily complex neural networks with weights $\theta_i$, which allows the NF to express complex, non-linear transformations. The inverse $\mB{u}=f_i^{-1}(\mB{w},\theta_i)$ of such a layer can be computed without having to invert $s(\cdot,\theta_i)$ or $t(\cdot,\theta_i)$ via
\begin{align}
\mB{u}_1&=&\text{exp}\big(-s(\mB{w}_2,\theta_i)\big)\odot\big(\mB{w}_1-t(\mB{w}_2,\theta_i)\big)\ \ \ \text{and}\ \ \ \mB{u}_2&=&\mB{w}_2\ .\nonumber
\end{align}
The Jacobian of an affine coupling layer is triangular and easy to compute (see \cite{Dinh_ICLR_2017} for details). In our model, we choose $s(\cdot,\theta_i)$ and $t(\cdot,\theta_i)$ to be fully-connected neural networks composed of $n_{\text{hid}}$ hidden layers and ReLU activations with shared weights $\theta_i$. As each coupling layer only affects half of the inputs, permuting or mixing them after each layer is crucial to allow for interaction between the dimensions. To do so, we simply reverse the order of the inputs after every second layer to make sure that all dimensions are able to contribute. The other coupling layers are linked by random, fixed orthogonal transformations that mix the data in an easily invertible way as proposed in \cite{Ardizzone_Arxiv_2019}.

The parameters $\theta=\{\theta_1,\ldots,\theta_{n_{\text{lay}}}\}$ (weights of all $n_{\text{lay}}$ scaling/translation functions) of our conditional NF model $f(\cdot;\theta)$ can be directly estimated using maximum likelihood training based on Eq. (\ref{eq:cov}). We choose a multivariate Gaussian distribution with diagonal unit covariance as a prior for $p(\mB{z})$ and also assume that the age prediction error $\|f_{a}(\mB{v};\theta)-a_{\text{gt}}\|_2$ with respect to ground-truth value $a_{\text{gt}}$ follows an univariate Gaussian distribution with small, user-defined covariance $\sigma^2$. This results in a negative log-likelihood loss term \cite{Kruse_ICML_2019,Ardizzone_Arxiv_2019,Xiao_Arxiv_2019}
\begin{equation}\label{eq:loss}
 \mathcal{L}(a,\mB{z})=\frac{1}{2}\Big(\sigma^{-2}\|a-a_{\text{gt}}\|_2^2+\|\mB{z}\|_2^2\Big)-\text{log}|J|\ .
\end{equation}
It is worth noting that this loss will only focus on correctly mapping velocity fields to the structured latent space (unidirectional training). By representing a bijective function, no two-way training is required to fit the bidirectional NF. Furthermore, replacing age $a$ with a vector would allow to build a conditional model that incorporates factors beyond age.

\section{Experiments and Results}\label{sec:results}
In our evaluation, we focus on showing that the conditional NF model for brain aging analysis derived in Sec. \ref{sec:problem} and Sec. \ref{sec:cnf} is able to (1) predict the biological brain age of previously unseen structural MR images, while (2) also serving as an age-conditioned, continuous generative model of brain morphology. To illustrate the second aspect, we use our NF model to generate conditional templates for different ages as, for example, also done in \cite{Dalca_NIPS_2019}.

\textbf{Data:} Two databases of T1-weighted brain MR images of 3730 healthy adults for which age data is available are used for our evaluation: (1) 3167 scans (age range: 20 -- 90 years) from the Study of Health in Pomerania (SHIP) \cite{Voelzke_SHIP_2011} for training (2684 randomly chosen subjects) and testing (483 subjects not used for training); (2) 563 scans (age range: 20 -- 86 years) from the publicly available IXI database\footnote{\url{https://brain-development.org/ixi-dataset/}} serve as an independent validation set.

For pre-processing (see Sec. \ref{sec:problem}), we start by computing a SHIP-specific brain template using all 49 scans of subjects younger than 26 years and the ANTs toolkit \cite{Avants_ANTs_2011}. We use young subjects here because they have little to no atrophy and a young template seems to be a reasonable starting point when the goal is to analyze and compare different aging effects/trajectories with respect to a single template. This 3D template (166$\times$209$\times$187 voxels; isotropic 1\,mm spacing) defines the reference space for our deformation-based modeling approach (see Sec. \ref{sec:problem:diffeo}). Subsequently, all 3730 scans are mapped to this brain template via non-linear diffeomorphic transformations (parameterized by stationary velocity fields) between the template brain and all subjects, which are computed using ITK's \textit{VariationalRegistration} module \cite{Werner_PMB_2014,Ehrhardt_BVM_2015}.

The velocity fields of the SHIP data are then used to estimate the affine subspace of maximum data variation via PCA to reduce the dimensionality of the modeling problem (see Sec. \ref{sec:problem:pca}). We choose $n_{\text{sub}}=500$ for our subspace, which covers $\approx 97\%$ of the variability of the training data.

\textbf{Experiments:}
Using the 2684 SHIP training subjects, we first train a conditional NF model as described in Sec. \ref{sec:nf:architecture} on a NVIDIA Quadro P4000 GPU with 8\,GB RAM and a TensorFlow 2.2 implementation. Our architecture consists of $n_{\text{lay}}=16$ affine coupling layers and each scaling/translation function is represented by a fully-connected neural network with $n_{\text{hid}}=2$ hidden layers of width $32$. Given an input size of $n_{\text{sub}}=500$, this setup results in $\approx 400$k trainable parameters $\theta$. Our batch size was selected to be equal to the total number of training samples and we optimize the loss function defined in Eq. (\ref{eq:loss}) with $\sigma=0.14$ for 20k epochs with an AdamW optimizer and a learning rate/weight decay of $10^{-4}$/$10^{-5}$. All parameters were chosen heuristically, but we found the results to be relatively insensitive to changes in $\sigma$ and $n_{\text{lay}}$.

\begin{table}
\caption{Mean absolute errors (MAEs) in years between the known chronological age and the predicted age obtained for the two approaches (MLR model \& NF model) when the trained models are applied to the SHIP test data subjects and all IXI subjects. Averaged results are reported for all subjects and six different age groups. See suppl. material for information about the number of subjects per age group.}
\label{tab:results}
\centering\footnotesize

\begin{tabular}{lcccccccc}
\footnotesize
\textbf{Model/}&\multicolumn{1}{@{}c@{\hskip 4.5\tabcolsep}}{\textbf{}}&\multicolumn{1}{@{}c@{\hskip 4.5\tabcolsep}}{\textbf{}}&\multicolumn{1}{@{}c@{\hskip 4.5\tabcolsep}}{\textbf{}}&\multicolumn{1}{@{}c@{\hskip 4.5\tabcolsep}}{\textbf{}}&\multicolumn{1}{@{}c@{\hskip 4.5\tabcolsep}}{\textbf{}}&\multicolumn{1}{@{}c@{\hskip 4.5\tabcolsep}}{}\\
\textbf{Age range}&\multicolumn{1}{@{}c@{\hskip 4.5\tabcolsep}}{$\mB{<40}$}&\multicolumn{1}{@{}c@{\hskip 4.5\tabcolsep}}{$\mB{40-50}$}&\multicolumn{1}{@{}c@{\hskip 4.5\tabcolsep}}{$\mB{50-60}$}&\multicolumn{1}{@{}c@{\hskip
4.5\tabcolsep}}{$\mB{60-70}$}&\multicolumn{1}{@{}c@{\hskip 4.5\tabcolsep}}{$\mB{70-80}$}&\multicolumn{1}{@{}c@{\hskip 4.5\tabcolsep}}{$\mB{>80}$}&\multicolumn{1}{@{}c@{\hskip 4.5\tabcolsep}}{\textbf{All}}\\
\midrule
&\multicolumn{7}{c}{\textit{SHIP data} (468 test subjects)}\\
MLR&\multicolumn{1}{@{}c@{\hskip 4.5\tabcolsep}}{5.34}&\multicolumn{1}{@{}c@{\hskip 4.5\tabcolsep}}{4.79}&\multicolumn{1}{@{}c@{\hskip 4.5\tabcolsep}}{5.33}&\multicolumn{1}{@{}c@{\hskip 4.5\tabcolsep}}{4.35}&\multicolumn{1}{@{}c@{\hskip 4.5\tabcolsep}}{6.13}&\multicolumn{1}{@{}c@{\hskip 4.5\tabcolsep}}{7.53}&\multicolumn{1}{@{}c@{\hskip 4.5\tabcolsep}}{\textbf{5.12}}\\
NF (ours)\ \ &\multicolumn{1}{@{}c@{\hskip 4.5\tabcolsep}}{5.35}&\multicolumn{1}{@{}c@{\hskip 4.5\tabcolsep}}{4.67}&\multicolumn{1}{@{}c@{\hskip 4.5\tabcolsep}}{5.10}&\multicolumn{1}{@{}c@{\hskip 4.5\tabcolsep}}{4.37}&\multicolumn{1}{@{}c@{\hskip 4.5\tabcolsep}}{6.29}&\multicolumn{1}{@{}c@{\hskip 4.5\tabcolsep}}{6.99}&\multicolumn{1}{@{}c@{\hskip 4.5\tabcolsep}}{\textbf{5.05}}\\
&\multicolumn{7}{c}{\textit{IXI data} (563 test subjects)}\\
MLR&\multicolumn{1}{@{}c@{\hskip 4.5\tabcolsep}}{6.54}&\multicolumn{1}{@{}c@{\hskip 4.5\tabcolsep}}{5.44}&\multicolumn{1}{@{}c@{\hskip 4.5\tabcolsep}}{6.19}&\multicolumn{1}{@{}c@{\hskip 4.5\tabcolsep}}{7.84}&\multicolumn{1}{@{}c@{\hskip 4.5\tabcolsep}}{7.71}&\multicolumn{1}{@{}c@{\hskip 4.5\tabcolsep}}{9.86}&\multicolumn{1}{@{}c@{\hskip 4.5\tabcolsep}}{\textbf{6.73}}\\
NF (ours)&\multicolumn{1}{@{}c@{\hskip 4.5\tabcolsep}}{8.47}&\multicolumn{1}{@{}c@{\hskip 4.5\tabcolsep}}{5.37}&\multicolumn{1}{@{}c@{\hskip 4.5\tabcolsep}}{4.92}&\multicolumn{1}{@{}c@{\hskip 4.5\tabcolsep}}{7.11}&\multicolumn{1}{@{}c@{\hskip 4.5\tabcolsep}}{6.57}&\multicolumn{1}{@{}c@{\hskip 4.5\tabcolsep}}{10.17}&\multicolumn{1}{@{}c@{\hskip 4.5\tabcolsep}}{\textbf{6.93}}\\
\bottomrule
\end{tabular}
\end{table}

The trained conditional NF model is then first used to predict the brain age of the SHIP test data subjects and all IXI subjects. Prediction accuracy is assessed by computing the mean absolute error (MAE) between the known chronological age and the predicted age. For comparison, we also train and evaluate the MAE of a multivariate linear regression (MLR) model on the same data to predict brain age from low-dimensional velocity field representations. Here, it is important to note that this MLR model also implicitly represents a non-linear map between brain morphology and age due to the non-linear relationship between deformation and velocity fields (group exponential map; see Sec. \ref{sec:problem:pca} and \cite{Arsigny_MICCAI_2006}).

Our conditional NF also provides an age-continuous generative model of brain aging. We show its capabilities by generating age-specific templates of the modeled population for different age values. Here, we assume that the conditional expectation $\mathbb{E}[\mathbf{v}|a]$ is an appropriate morphology template for a given age $a$ and compute velocity vectors $\mathbf{v}$ for ages $a=\{40,50,60,70,80,90\}$ via Monte-Carlo approximation with 150k random samples. This is done by fixing the age component of the latent space and sampling from the NF's prior for the variability part. The vectors are then mapped back to the high-dimensional space of velocity fields using $\mathbf{Q}^T$ (see Sec. \ref{sec:problem:pca}) to obtain the associated diffeomorphic transformations. These transformations are then applied to the SHIP-specific brain template to generate structural MR images for the age-conditioned morphology templates.

\begin{figure}
\centering
\includegraphics[width=0.9\textwidth]{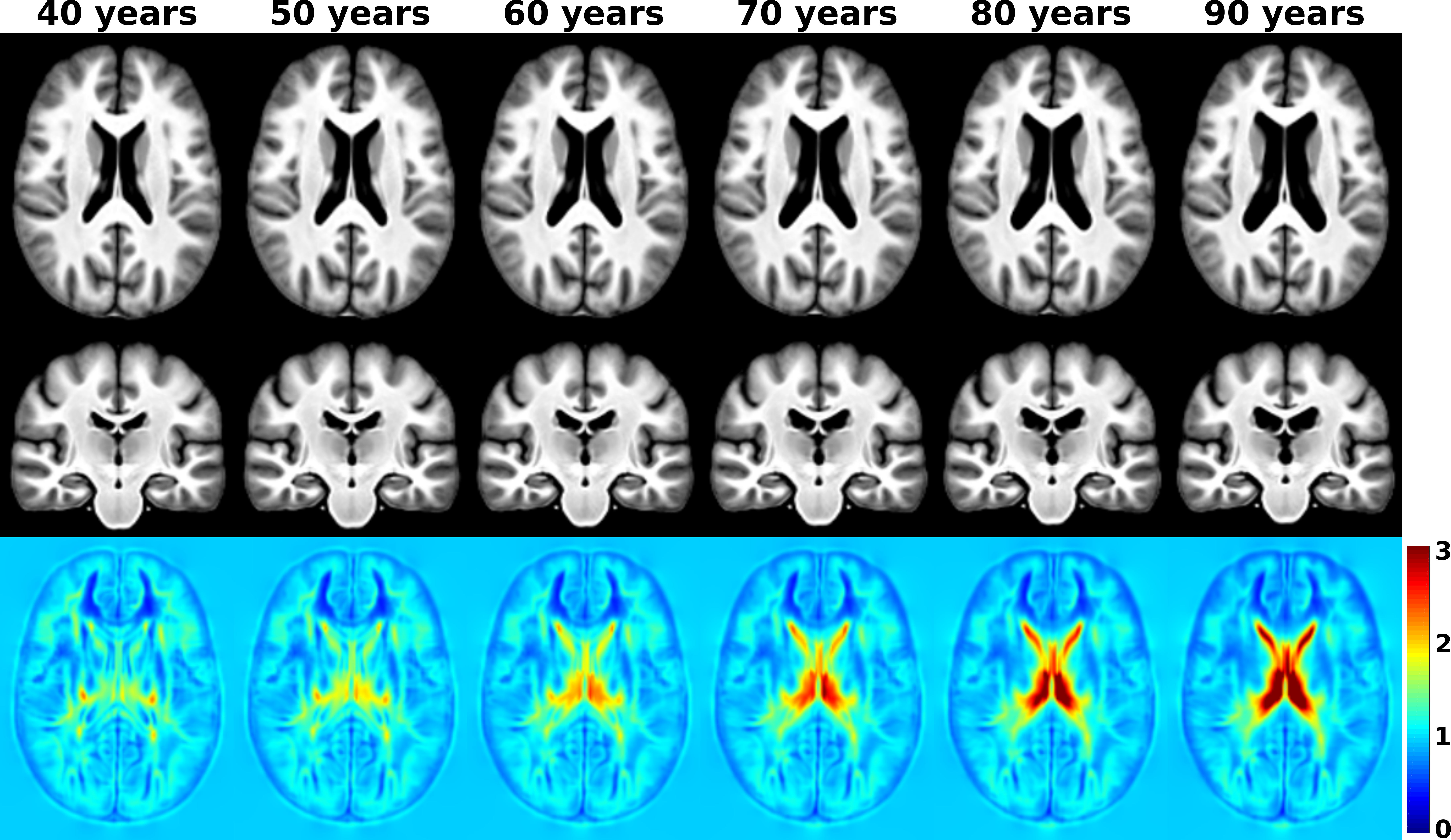}
\caption{Selected slices of age-specific morphology templates generated using our NF model for different ages. Last row: Corresponding axial slices of the Jacobian determinants for the generated transformations with respect to the (young) reference template, which clearly show the increase in ventricle size (values $>1$) and the shrinking trend in other areas (values $< 1$).}\label{fig:results}
\end{figure}

\textbf{Results:} The evaluation results are summarized in Tab. \ref{tab:results} and Fig. \ref{fig:results}. Our NF model achieves an overall MAE of $5.05$ years for the SHIP data and a MAE of $6.93$ years for the IXI data, respectively. Both results are comparable to the overall accuracy of the MLR model (insignificant differences; paired t-test with $p=0.79$/$p=0.13$). However, their performance differs for certain age groups. Interestingly, for the IXI data,  which is known to be a challenging dataset due to its variability \cite{Ethan_Cortex_2019}, the NF clearly outperforms the MLR model for subjects between 50 and 80 years. We believe that this finding indicates that our NF model is able to better capture the general non-linear trend of the aging process. 

The conditional templates for different ages displayed in Fig. \ref{fig:results} illustrate that our NF model is able to capture the typical trend of healthy brain aging. For example, the total ventricle volume increases by a factor of 2.02 between age 40 and age 90, while the total putamen volume decreases by a factor of 0.87. Volumes changes were quantified based on segmentations propagated from the template. Furthermore, the general shape characteristics of the different templates are stable across the age range, which indicates that the NF disentangles aging and non-aging factors (see also Fig. 1 of the suppl. material).

\section{Conclusion}\label{sec:discuss}
In this paper, we propose a new method to model bidirectional functional relationships between 3D brain morphology and brain age in a unified way using conditional normalizing flows. In an initial evaluation, we showed that our unified model can accurately predict biological brain age while also being able to generate age-conditioned brain templates. Based on the evaluation results, our future work will primarily focus on obtaining more training data and improving our architecture to obtain even more accurate age prediction results. We also plan to compare its performance to different prediction and generative modeling approaches from the literature and to condition the model on additional factors beyond age.
\subsubsection{Acknowledgements}
This work was supported by the University of Calgary's Eyes High postdoctoral scholarship program and the River Fund at Calgary Foundation.
%
%
\bibliographystyle{splncs04}

\newpage
\section*{Supplementary Material}
\begin{figure}[h]
  \centering
  \includegraphics[width=.15\textwidth]{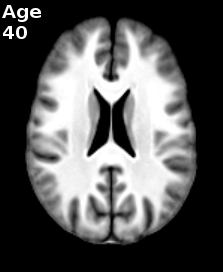}  
  \includegraphics[width=.15\textwidth]{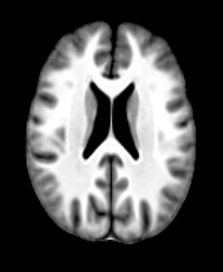}  
  \includegraphics[width=.15\textwidth]{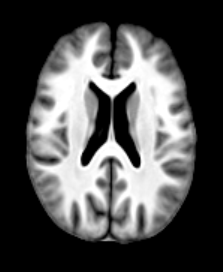}
  \includegraphics[width=.15\textwidth]{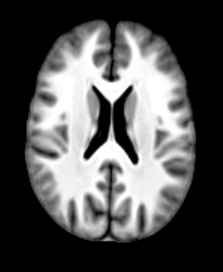}
  \includegraphics[width=.15\textwidth]{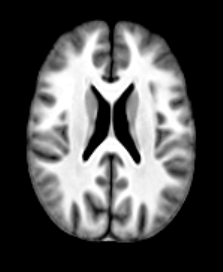}
  \includegraphics[width=.15\textwidth]{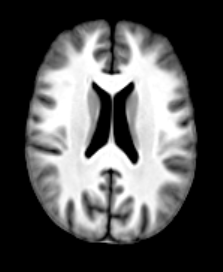}
  \includegraphics[width=.15\textwidth]{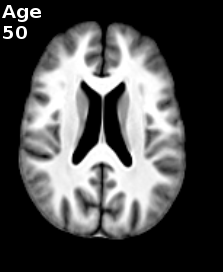}
  \includegraphics[width=.15\textwidth]{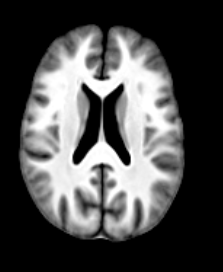}
  \includegraphics[width=.15\textwidth]{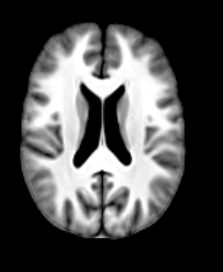}
  \includegraphics[width=.15\textwidth]{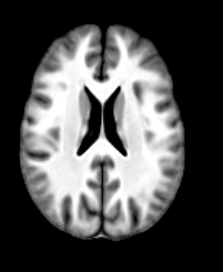}
  \includegraphics[width=.15\textwidth]{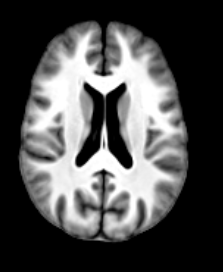}
  \includegraphics[width=.15\textwidth]{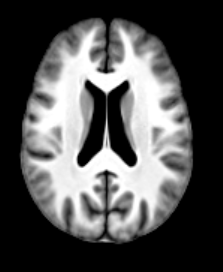}
  \includegraphics[width=.15\textwidth]{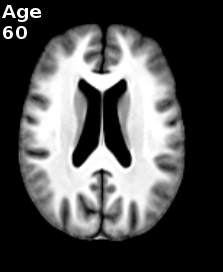}
  \includegraphics[width=.15\textwidth]{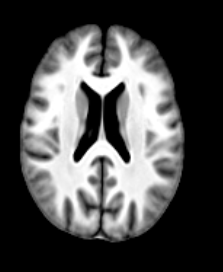}
  \includegraphics[width=.15\textwidth]{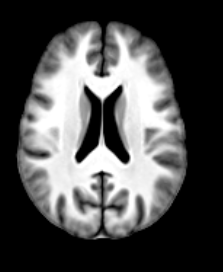}
  \includegraphics[width=.15\textwidth]{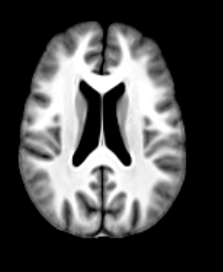}
  \includegraphics[width=.15\textwidth]{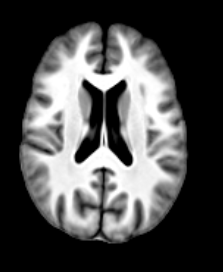}
  \includegraphics[width=.15\textwidth]{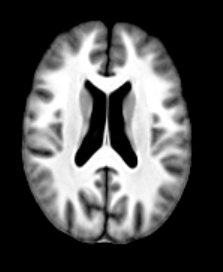}
  \includegraphics[width=.15\textwidth]{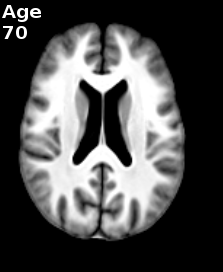}
  \includegraphics[width=.15\textwidth]{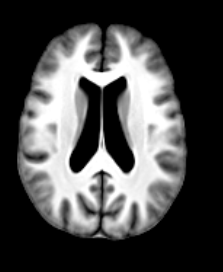}
  \includegraphics[width=.15\textwidth]{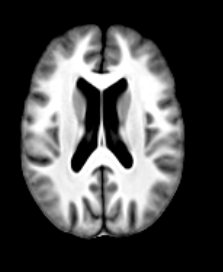}
  \includegraphics[width=.15\textwidth]{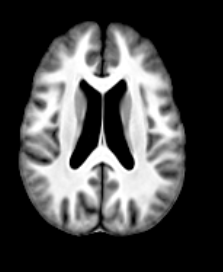}
  \includegraphics[width=.15\textwidth]{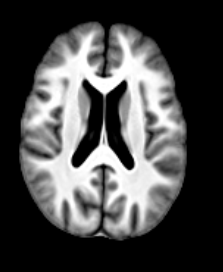}
  \includegraphics[width=.15\textwidth]{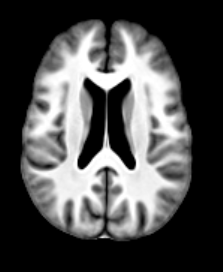}
  \includegraphics[width=.15\textwidth]{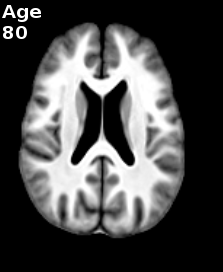}
  \includegraphics[width=.15\textwidth]{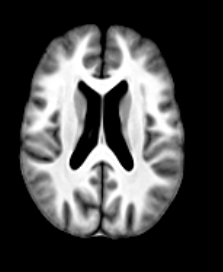}
  \includegraphics[width=.15\textwidth]{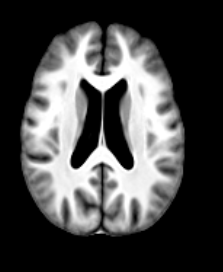}
  \includegraphics[width=.15\textwidth]{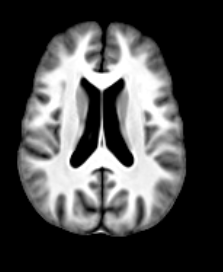}
  \includegraphics[width=.15\textwidth]{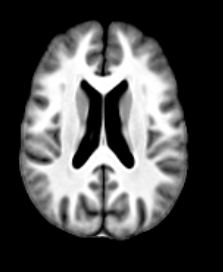}
  \includegraphics[width=.15\textwidth]{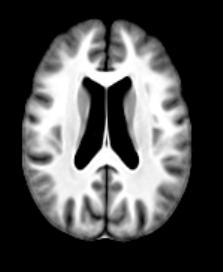}
  \includegraphics[width=.15\textwidth]{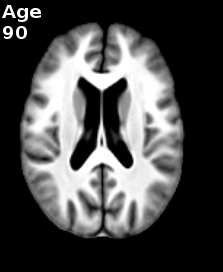}
  \includegraphics[width=.15\textwidth]{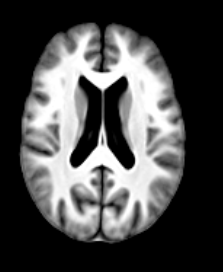}
  \includegraphics[width=.15\textwidth]{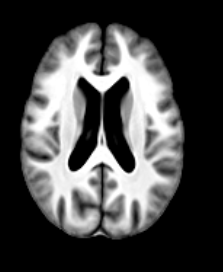}
  \includegraphics[width=.15\textwidth]{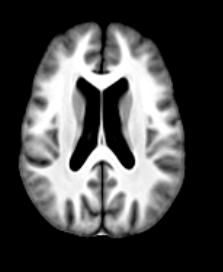}
  \includegraphics[width=.15\textwidth]{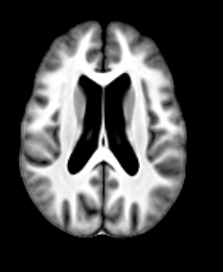}
  \includegraphics[width=.15\textwidth]{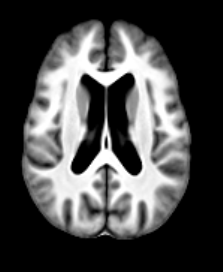}
\caption{Selected slices of random brain morphology samples generated using our NF model for different ages showing the variability captured by the model. Rows are sorted by age in an increasing order (10 years difference between each row) from age 40 (top row) to age 90 (bottom row). The general trend of healthy brain aging is clearly visible (e.g., larger ventricles with increasing age), while each row also shows that diverse samples can be generated for each specific age.}
\end{figure}

\begin{table}
\caption{Number of subjects per age group for both databases used. For the SHIP data, numbers of training/test subjects are reported separately while IXI subjects were only used for testing.}
\label{tab:results}
\centering\footnotesize

\begin{tabular}{lcccccccc}
\footnotesize
\textbf{Data/}&\multicolumn{1}{@{}c@{\hskip 4.5\tabcolsep}}{\textbf{}}&\multicolumn{1}{@{}c@{\hskip 4.5\tabcolsep}}{\textbf{}}&\multicolumn{1}{@{}c@{\hskip 4.5\tabcolsep}}{\textbf{}}&\multicolumn{1}{@{}c@{\hskip 4.5\tabcolsep}}{\textbf{}}&\multicolumn{1}{@{}c@{\hskip 4.5\tabcolsep}}{\textbf{}}&\multicolumn{1}{@{}c@{\hskip 4.5\tabcolsep}}{}\\
\textbf{Age range}&\multicolumn{1}{@{}c@{\hskip 4.5\tabcolsep}}{$\mB{<40}$}&\multicolumn{1}{@{}c@{\hskip 4.5\tabcolsep}}{$\mB{40-50}$}&\multicolumn{1}{@{}c@{\hskip 4.5\tabcolsep}}{$\mB{50-60}$}&\multicolumn{1}{@{}c@{\hskip
4.5\tabcolsep}}{$\mB{60-70}$}&\multicolumn{1}{@{}c@{\hskip 4.5\tabcolsep}}{$\mB{70-80}$}&\multicolumn{1}{@{}c@{\hskip 4.5\tabcolsep}}{$\mB{>80}$}&\multicolumn{1}{@{}c@{\hskip 4.5\tabcolsep}}{\textbf{All}}\\
\midrule
&\multicolumn{7}{c}{\# training subjects/\# test subjects}\\
SHIP data&\multicolumn{1}{@{}c@{\hskip 4.5\tabcolsep}}{482/86}&\multicolumn{1}{@{}c@{\hskip 4.5\tabcolsep}}{663/114}&\multicolumn{1}{@{}c@{\hskip 4.5\tabcolsep}}{634/119}&\multicolumn{1}{@{}c@{\hskip 4.5\tabcolsep}}{579/103}&\multicolumn{1}{@{}c@{\hskip 4.5\tabcolsep}}{296/55}&\multicolumn{1}{@{}c@{\hskip 4.5\tabcolsep}}{30/6}&\multicolumn{1}{@{}c@{\hskip 4.5\tabcolsep}}{\textbf{2684/483}}\\
IXI data\ \ &\multicolumn{1}{@{}c@{\hskip 4.5\tabcolsep}}{200}&\multicolumn{1}{@{}c@{\hskip 4.5\tabcolsep}}{89}&\multicolumn{1}{@{}c@{\hskip 4.5\tabcolsep}}{99}&\multicolumn{1}{@{}c@{\hskip 4.5\tabcolsep}}{118}&\multicolumn{1}{@{}c@{\hskip 4.5\tabcolsep}}{49}&\multicolumn{1}{@{}c@{\hskip 4.5\tabcolsep}}{8}&\multicolumn{1}{@{}c@{\hskip 4.5\tabcolsep}}{\textbf{563}}\\
\bottomrule
\end{tabular}
\end{table}

\end{document}